\documentclass{article}

\usepackage[utf8]{inputenc}

\usepackage[margin=1in,top=1in,bottom=1in,left=1in,right=1in,includehead,includefoot]{geometry}

\usepackage{amsthm}
\usepackage{thmtools}

\usepackage{preamble}

\usepackage{titlesec}

\usepackage{soul}
\usepackage{xcolor}
\sethlcolor{yellow}

\usepackage{verbatim}
\usepackage{booktabs}
\usepackage{amsmath}
\usepackage{mathtools}
\usepackage{amssymb}
\usepackage{tabularx}
\usepackage{url}
\usepackage{hyperref}
\usepackage{setspace}
\usepackage{bm}
\usepackage[justification=centering]{caption}
\usepackage{float}
\usepackage{algorithm,algorithmicx,algpseudocode}
\usepackage{graphicx}

\bibliography{main}

\title{Interpretable and Adaptive Node Classification on Heterophilic Graphs via Combinatorial Scoring and Hybrid Learning}
\author{
  Soroush Vahidi\thanks{Department of Computer Science, New Jersey Institute of Technology, Newark, NJ, USA. \texttt{sv96@njit.edu}} 
}
\date{}

\begin{document}

\maketitle
\begin{abstract}
Graph neural networks (GNNs) achieve strong performance on homophilic graphs but often struggle under heterophily, where adjacent nodes frequently belong to different classes. We propose an interpretable and adaptive framework for semi-supervised node classification based on explicit combinatorial inference rather than deep message passing. Our method assigns labels using a confidence-ordered greedy procedure driven by an additive scoring function that integrates class priors, neighborhood statistics, feature similarity, and training-derived label–label compatibility. A small set of transparent hyperparameters controls the relative influence of these components, enabling smooth adaptation between homophilic and heterophilic regimes.

We further introduce a validation-gated hybrid strategy in which combinatorial predictions are optionally injected as priors into a lightweight neural model. Hybrid refinement is applied only when it improves validation performance, preserving interpretability when neuralization is unnecessary. All adaptation signals are computed strictly from training data, ensuring a leakage-free evaluation protocol. Experiments on heterophilic and transitional benchmarks demonstrate competitive performance with modern GNNs while offering advantages in interpretability, tunability, and computational efficiency.
\end{abstract}

\section{Introduction}

\subsection{Motivation: Toward General Node Classification Across Graph Structures}

Node classification is a central problem in graph-based machine learning, with applications spanning fraud detection, social and information networks, recommendation systems, and biological analysis~\cite{MR24,MB21}. Over the past decade, Graph Neural Networks (GNNs) have emerged as the dominant modeling paradigm for this task, achieving strong performance on benchmarks characterized by \emph{homophily}, where adjacent nodes are likely to share the same label. These successes rely heavily on message-passing mechanisms that aggregate neighborhood information under an implicit assumption of label similarity~\cite{GCN17,SGC19,APPNP22}.

However, many real-world graphs violate this assumption. In \emph{heterophilic} networks, edges frequently connect nodes of different classes, causing standard message-passing architectures to propagate misleading signals. To address this mismatch, a growing body of work has proposed heterophily-aware neural architectures, including H2GCN~\cite{H2GCN22}, FAGCN~\cite{FAGCN21}, GPR-GNN~\cite{GPRGNN}, and more recently Diffusion-Jump GNN (DJ-GNN)~\cite{DJGNN25}, which leverages spectral learning and high-order diffusion filters to achieve state-of-the-art accuracy on several challenging benchmarks.

While these neural models substantially improve performance under heterophily, they often introduce increased architectural complexity, substantial computational overhead, and limited interpretability. Moreover, their inductive biases are typically implicit, making it difficult to diagnose failures or to reason about how structural signals influence predictions. This motivates the question of whether node classification across diverse graph regimes can be addressed using a simpler and more transparent inference paradigm.

In this work, we propose an interpretable and general-purpose \emph{combinatorial} framework for semi-supervised node classification that departs from deep message passing. Instead of learning latent representations, our method assigns labels through an explicit additive scoring function that integrates class priors, neighborhood statistics, feature similarity, and training-derived label--label compatibility. A small set of interpretable hyperparameters governs the relative influence of these components, enabling direct control over the model’s inductive bias. In particular, the sign and magnitude of a key parameter (\(a_2\)) explicitly determine whether neighborhood evidence is treated as supportive (homophilic) or adversarial (heterophilic), allowing the method to interpolate smoothly between structural regimes.

Although the proposed approach does not aim to universally outperform specialized neural architectures such as DJ-GNN in terms of raw accuracy, it offers complementary advantages that are often critical in practice. These include strong interpretability, lightweight inference, and fine-grained user control over the classification process. Once hyperparameters are selected, inference is extremely fast, and the contribution of each evidence source to a prediction can be inspected directly.

To further enhance adaptability without sacrificing transparency, we introduce a validation-gated hybrid strategy that conditionally combines the combinatorial predictor with a lightweight neural model based on the ALT framework~\cite{xc23}. Combinatorial predictions are used as priors for neural refinement, but hybridization is applied only when it improves validation performance. This conservative decision rule treats neural modeling as an optional refinement rather than a mandatory component, ensuring that additional complexity is introduced only when empirically justified.

Together, these design choices position our framework as a robust and interpretable alternative to deep message-passing architectures—particularly well suited for settings where graph structure varies across regimes, computational efficiency matters, or transparent decision-making is required.

\subsection{Contributions}

Recent heterophily-aware graph neural networks, such as Diffusion-Jump GNN (DJ-GNN)~\cite{DJGNN25}, achieve strong predictive performance across diverse graph regimes, but often do so at the cost of architectural complexity, limited interpretability, and substantial computational overhead. In contrast, we advocate a complementary modeling paradigm based on explicit combinatorial inference, designed to provide transparency, controllability, and efficiency without relying on deep message passing.

Our main contributions are summarized as follows:

\begin{itemize}
    \item \textbf{Interpretable combinatorial node classification.}  
    We introduce a neural-free node classification algorithm that assigns labels via an explicit additive scoring rule integrating class priors, neighborhood statistics, feature similarity, and training-derived label--label compatibility. Each component contributes transparently to the final decision, enabling direct inspection and diagnosis of predictions.

    \item \textbf{Explicit adaptation across homophilic and heterophilic regimes.}  
    The proposed framework exposes a small set of interpretable hyperparameters that directly control how structural and feature-based evidence is used. In particular, the parameter \(a_2\) governs the role of neighboring labels, promoting homophilic behavior when positive and heterophilic behavior when negative, allowing the model to interpolate smoothly between graph regimes.

    \item \textbf{Validation-gated hybrid combinatorial--neural refinement.}  
    We propose a conservative hybrid strategy in which combinatorial predictions are optionally injected as soft priors into a lightweight neural model based on the ALT framework~\cite{xc23}. Neural refinement is activated only when it improves validation performance, providing a principled trade-off between interpretability and predictive flexibility.

    \item \textbf{Strong diagnostic and interpretability properties.}  
    Unlike black-box neural models, the proposed approach provides transparent control over label propagation, neighborhood influence, and feature similarity. Hyperparameters have clear semantic meaning, making the method easy to tune, analyze, and adapt to domain-specific requirements.

    \item \textbf{Competitive performance with lightweight computation.}  
    Extensive experiments on heterophilic, transitional, and homophilic benchmarks demonstrate that the proposed framework achieves competitive accuracy relative to modern neural baselines, while offering substantially lower inference cost and favorable behavior in resource-constrained or interpretability-critical settings.
\end{itemize}

Together, these contributions establish an interpretable and adaptive alternative to deep message-passing architectures, positioning combinatorial inference as a viable and practical foundation for node classification across diverse graph structures.

\subsection{Summary of Results}

We evaluate the proposed framework on widely used node classification benchmarks
spanning strongly heterophilic, weak-signal, transitional, and homophilic graph
regimes. Strongly heterophilic settings are represented by the WebKB datasets
\textbf{Texas}, \textbf{Cornell}, and \textbf{Wisconsin}. The \textbf{Actor}
dataset constitutes a challenging non-homophilic benchmark with weak and noisy
structural signal. \textbf{CiteSeer} serves as a transitional regime, while
\textbf{Cora} and \textbf{Pubmed} represent strongly homophilic citation networks.

On the heterophilic WebKB datasets, the proposed combinatorial predictor achieves
competitive accuracy and consistently outperforms classical message-passing GNNs,
including GCN~\cite{GCN17}, SGC~\cite{SGC19}, and
APPNP~\cite{APPNP22}. While our approach does not uniformly surpass specialized
heterophily-aware architectures such as DJ-GNN~\cite{DJGNN25} in terms of raw
accuracy, it provides clear advantages in interpretability, tunability, and
computational efficiency, particularly under strong heterophily.

The Actor dataset highlights a complementary regime in which neighborhood
aggregation provides limited benefit due to weak alignment between graph
structure and class labels. Consistent with prior work, all evaluated methods
exhibit a low empirical performance ceiling on Actor. In this setting, the
validation gate in our framework frequently selects the standalone combinatorial
predictor, reflecting the limited utility of neural refinement and preserving
robust, interpretable behavior.

We further examine a hybrid variant of the framework in which combinatorial
predictions are used as priors for a lightweight neural model based on the ALT
framework~\cite{xc23}. Hybrid refinement can yield additional improvements in
settings where feature propagation or structural learning provides complementary
information, such as Wisconsin and transitional citation networks. Because the
benefit of refinement varies across datasets and splits, we employ a
validation-gated decision rule that dynamically selects between the standalone
combinatorial predictor and the hybrid model. This mechanism ensures that neural
refinement is applied only when it improves validation performance, preserving
transparency and robustness when the combinatorial predictor alone is sufficient.

\subsection{Related Work}

\paragraph{Feature-only and message-passing baselines.}
Multi-layer perceptrons (MLPs) are commonly used as feature-only baselines for
node classification, operating solely on node attributes while ignoring graph
structure~\cite{h18}. Most graph neural networks (GNNs) extend this paradigm by
incorporating neighborhood aggregation to compute latent node representations,
followed by an MLP-style classifier. A foundational model in this class is the
Graph Convolutional Network (GCN)~\cite{GCN17}, which introduced an efficient
layer-wise propagation rule based on normalized adjacency matrices and achieved
strong performance on homophilic benchmarks. Several simplified variants were
later proposed, including Simple Graph Convolution (SGC)~\cite{SGC19}, which
removes nonlinearities and collapses multiple layers into a single linear
transformation, substantially reducing complexity while retaining competitive
accuracy.

GraphSAGE is an inductive representation learning framework that generates node
embeddings by sampling and aggregating feature information from local
neighborhoods, enabling generalization to unseen nodes and even entirely new
graphs~\cite{h18}. By learning parameterized aggregation functions rather than
node-specific embeddings, GraphSAGE scales efficiently to large, evolving
graphs. GCNI extends standard GCNs to deeper architectures through the
introduction of initial residual connections and identity mappings, which
mitigate over-smoothing and preserve input feature information across many
layers, enabling effective training at increased depth~\cite{GCNI}.  
Geom-GCN introduces a geometric aggregation framework that embeds nodes into a
latent continuous space and performs permutation-invariant bi-level aggregation
over graph-based and geometry-induced neighborhoods. This design explicitly
captures structural information and enables access to informative long-range
interactions, leading to strong performance on both homophilic and heterophilic
benchmarks where standard message-passing GNNs are limited~\cite{p20}.

\paragraph{Propagation-based and spectral adaptations.}
To mitigate the limited receptive field and oversmoothing effects of shallow
GCNs, APPNP~\cite{APPNP22} decouples prediction from propagation by applying a
personalized PageRank diffusion over initial node predictions. GPR-GNN
\cite{GPRGNN} generalizes this idea by learning adaptive propagation weights,
allowing interpolation between local and long-range aggregation and improved
robustness across homophilic and heterophilic regimes. Related efforts such as
FAGCN~\cite{FAGCN21} explicitly incorporate both low- and high-frequency graph
signals through frequency-adaptive filtering, improving stability under
disassortative structure.

\paragraph{Structure--feature decoupling and convolutional baselines.}
LINKX is a simple yet effective model for learning on non-homophilous graphs that
separately embeds node features and graph structure using MLPs, and then combines
them through lightweight transformations rather than message passing~\cite{l20}.
By decoupling adjacency and feature information, LINKX avoids oversmoothing and
scales efficiently, making it a strong baseline in heterophilic settings. CGCN is
a convolution-based graph neural network that applies localized convolution
operators directly on graph neighborhoods, aiming to generalize classical
convolutional ideas to irregular graph structures~\cite{cgcn}. While conceptually
simple, CGCN relies on fixed aggregation schemes and is less flexible in handling
strong heterophily.

\paragraph{Architectural designs for heterophily.}
Several models explicitly target heterophilic graphs through architectural
choices. H2GCN~\cite{H2GCN22} identifies key design principles, including the
separation of ego- and neighbor-embeddings and higher-order aggregation, that
improve performance when neighbors tend to have different labels. Ordered GNN
(OGNN)~\cite{OGNN23} introduces an ordered message-passing mechanism that aligns
multi-hop information flow with structured embedding dimensions, addressing both
heterophily and oversmoothing. ASGC~\cite{ASGC22} extends simplified graph
convolutions by learning adaptive spectral filters via least-squares regression,
retaining interpretability while supporting non-smoothing behavior. FSGNN
\cite{FSGNN21} further decouples feature propagation from representation learning
by independently transforming multi-hop features and softly selecting among
them, achieving strong results with a shallow architecture.

\paragraph{Attention, normalization, and structure learning.}
Graph Attention Networks (GAT)~\cite{GAT18} introduce attention-weighted
aggregation, allowing nodes to assign different importances to their neighbors
during message passing. While effective on several benchmarks, GAT remains a
fully neural aggregation model and can be sensitive to heterophilic structure.
PairNorm~\cite{PairNorm19} addresses oversmoothing through normalization that
preserves pairwise feature distances across layers, improving stability in deep
GNNs without modifying the aggregation operator. Beyond architectural
modifications, the ALT framework~\cite{xc23} proposes a structure-learning
approach that adaptively reweights and integrates multiple graph signals,
achieving strong performance across varying homophily levels.

\paragraph{Diffusion-based heterophily modeling.}
Diffusion-Jump GNN (DJ-GNN)~\cite{DJGNN25} models structural heterophily through
diffusion geometry, learning spectral metrics that guide aggregation over distant
but semantically related nodes. By combining diffusion distances with
classification objectives, DJ-GNN achieves state-of-the-art performance on
heterophilic benchmarks and represents one of the strongest neural baselines in
this regime.

\paragraph{Positioning of our work.}
In contrast to prior approaches that primarily rely on architectural depth,
spectral filtering, or learned message passing, our method adopts an explicit
combinatorial inference rule with transparent scoring components and
validation-gated hybridization. This design emphasizes interpretability,
controllability, and adaptive behavior across homophilic and heterophilic
settings without introducing deep or opaque architectures.

\section{Interpretable and Adaptive Combinatorial--Neural Node Classification}

\subsection{Problem Setting and Design Principles}
\label{sec:method:setting}

\paragraph{Semi-supervised node classification.}
Let $\mathcal{G}=(V,E,X,y)$ be a graph with $|V|=n$ nodes, edges $E$, node features
$X\in\mathbb{R}^{n\times d}$, and discrete labels $y\in\mathcal{Y}^n$, where only a
subset of nodes is labeled for training. We consider the standard transductive
semi-supervised setting: given a training index set $T\subseteq V$, we observe
$\{(X_v,y_v): v\in T\}$ and aim to predict labels on a disjoint target set,
typically a validation set $V_{\text{val}}$ and a test set $V_{\text{test}}$.
Throughout, evaluation is performed exclusively on held-out nodes, and all
hyperparameter selection is carried out using \emph{only training nodes}, with
cross-validation splits drawn solely from $T$, to avoid any form of information
leakage.

\paragraph{Homophily, heterophily, and inductive bias.}
Many message-passing graph neural networks rely on an implicit \emph{homophily}
assumption, whereby adjacent nodes are likely to share labels. Under this
assumption, neighborhood aggregation provides a strong inductive bias. In
\emph{heterophilic} graphs, however, edges frequently connect nodes of different
classes, and the same aggregation mechanism may become unreliable. Our objective
is therefore to design a method that (i) remains effective across a broad spectrum
of homophily and heterophily regimes, (ii) makes its structural assumptions
explicit through decomposable scoring components, and (iii) avoids reliance on
deep or opaque message-passing architectures.

\paragraph{Core design principles.}
Our framework is guided by the following principles:

\begin{enumerate}
  \item \textbf{Decomposable additive scoring.}
  Predictions are produced by an explicit additive scoring rule whose components
  correspond to distinct sources of evidence, including global class priors,
  neighborhood label statistics, feature similarity to class prototypes, and
  train-derived label--label compatibility. For any node--class pair, the
  numerical contribution of each component can be inspected independently,
  providing a transparent accounting of how different evidence sources influence
  the final decision.

  \item \textbf{Confidence-ordered inference.}
  Rather than labeling all unlabeled nodes simultaneously, nodes are processed in
  an order determined by an explicit \emph{priority} function reflecting the
  reliability of available evidence, such as the number of labeled neighbors and
  feature-based confidence. This greedy and deterministic procedure favors nodes
  with stronger evidence early in inference and improves reproducibility.

  \item \textbf{Explicit adaptation parameters.}
  Instead of hard-coding a homophilic or heterophilic assumption, the method
  exposes a small set of interpretable hyperparameters that control the relative
  influence of neighbor agreement, feature similarity, and label--label
  compatibility. These parameters allow the scoring rule to interpolate smoothly
  between homophilic and heterophilic regimes and are tuned using training data
  only.

  \item \textbf{Hybrid refinement only when beneficial.}
  A lightweight neural refinement stage can optionally be applied, but it is
  treated as a validation-gated extension rather than a mandatory component.
  Specifically, we compare the standalone combinatorial predictor against a
  hybrid model on the validation set, where combinatorial predictions are injected
  as logit-level priors. The strategy achieving higher validation accuracy is
  selected for final test evaluation.
\end{enumerate}

\paragraph{Method overview.}
We summarize the framework executed for each data split as follows.
Hyperparameters of the combinatorial predictor are tuned using a robust random
search procedure with cross-validation performed exclusively on the training
set. The resulting combinatorial predictor serves as a decomposable baseline
whose scoring components can be inspected directly. Optionally, its predictions
are used as priors in a lightweight neural model. A validation-based gating
mechanism then selects between the baseline and hybrid strategies for final
evaluation. The remainder of the Method section details the combinatorial
inference rule (Section~\ref{sec:method:combinatorial}), the mechanisms enabling
adaptation to homophilic and heterophilic graph structure
(Section~\ref{sec:method:adaptation}), and the validation-gated hybrid refinement
strategy (Section~\ref{sec:method:hybrid}).

\subsection{Combinatorial Scoring and Confidence-Ordered Label Propagation}
\label{sec:method:combinatorial}

\paragraph{Overview.}
Given a training set $T$ and a target set $U$ (validation or test), the combinatorial
predictor assigns labels to nodes in $U$ using a confidence-ordered, greedy
procedure. The method maintains (i) incremental counts of labeled neighbors per
class, (ii) class prototypes in feature space, and (iii) a label--label
compatibility matrix estimated from training data. At each step, a single
unlabeled node is selected according to a priority function and assigned the
class maximizing an explicit additive score. Predictions for nodes outside $U$
are not used for evaluation.

\paragraph{Graph preprocessing.}
For inference, the input graph is converted into an undirected simple graph by
symmetrizing the edge set and removing duplicate edges, yielding an undirected
edge index $\widetilde{E}$. All neighborhood statistics used by the combinatorial
predictor are computed on $\widetilde{E}$, ensuring consistent treatment of
directed and undirected datasets.

\paragraph{Training-derived priors and compatibility.}
Let $\mathcal{Y}_{\text{used}}$ denote the set of labels present in the dataset and
$C = |\mathcal{Y}_{\text{used}}|$. A Laplace-smoothed class prior is computed from
the training labels:
\begin{equation}
\pi_c = \frac{n_c + \alpha}{|T| + \alpha C},
\end{equation}
where $n_c$ is the number of training nodes of class $c$ and $\alpha > 0$.

To capture structural interactions between labels, a label--label compatibility
matrix is estimated using \emph{only} edges whose endpoints both lie in $T$. For
each undirected training edge $(u,v) \in \widetilde{E}$, the ordered pair
$(y_v, y_u)$ is counted and normalized row-wise with Laplace smoothing:
\begin{equation}
\mathrm{Compat}[c_n, c] \approx \Pr(y_u = c \mid y_v = c_n).
\end{equation}
When training--training edges are sparse, smoothing and adaptive reweighting
naturally limit the influence of compatibility-based evidence.

\paragraph{Feature prototypes and similarity.}
For each class $c$, the predictor maintains a prototype vector $\mu_c$ computed
as a weighted centroid in feature space. Training-labeled nodes contribute weight
$(1-a_7)$, while nodes labeled during propagation contribute weight $a_7$,
allowing propagated labels to influence prototypes in a controlled manner.
Feature agreement is measured using cosine similarity,
rescaled to $[0,1]$:
\begin{equation}
D(u,c) = \frac{1 + \cos(x_u, \mu_c)}{2}.
\end{equation}
Prototypes and similarities are periodically recomputed to reflect newly labeled
nodes.

\paragraph{Neighborhood statistics and compatibility support.}
Let $\mathrm{cnt}(u,c)$ denote the number of labeled neighbors of node $u$
currently assigned to class $c$, and let $\deg(u)$ be the degree of $u$ in
$\widetilde{E}$. We define the degree-normalized neighbor distribution
\begin{equation}
p_u(c) = \frac{\mathrm{cnt}(u,c)}{\deg(u)}.
\end{equation}
When labeled neighbors are available, a compatibility support term is computed as
\begin{equation}
s_u(c) = \sum_{c_n} w_u(c_n)\,\mathrm{Compat}[c_n, c],
\end{equation}
where $w_u(c_n)$ is the normalized distribution of labeled neighbor classes.
When few labeled neighbors are present, the influence of neighborhood and
compatibility evidence is adaptively attenuated to reduce unreliable early
propagation.

\paragraph{Additive scoring rule.}
Each unlabeled node $u \in U$ is assigned the label maximizing the explicit
additive score
\begin{equation}
\mathrm{Score}_u(c)
= a_1\,\pi_c
+ a_2\,p_u(c)
+ a_3\,D(u,c)
+ a_8\,s_u(c),
\end{equation}
where the coefficients $a_1, a_2, a_3, a_8$ control the relative influence of
global priors, direct neighbor agreement, feature similarity, and
compatibility-based structural support. These terms correspond to distinct and
independently inspectable sources of evidence for each node--class pair.

\paragraph{Priority function and confidence-ordered inference.}
Unlabeled nodes are processed in descending order of a priority function that
combines (i) the number of labeled neighbors originating from $T$, (ii) the
number of already propagated neighbors, both normalized by degree and weighted
separately, and (iii) a feature-based confidence term proportional to
$\max_c D(u,c)$. This ordering prioritizes nodes with stronger and more reliable
evidence early in inference.

To reduce error amplification, the influence of neighborhood- and
compatibility-based terms is dynamically scaled based on the number of labeled
neighbors. Optionally, nodes whose best and second-best scores are insufficiently
separated may be deferred and reconsidered later in the inference process.

\paragraph{Refresh schedule and determinism.}
After labeling a fixed fraction of nodes (every
$\max(\lfloor n/5\rfloor,1)$ newly labeled nodes), class prototypes, feature
similarities, and node priorities are recomputed, and the priority queue is
rebuilt. When multiple classes achieve near-identical scores, ties are broken
deterministically using a fixed lexicographic rule based on neighborhood
agreement, feature similarity, class prior, and label index, ensuring
reproducible inference.

\paragraph{Algorithm.}
The combinatorial predictor is implemented as \texttt{predictclass} in our code
and is fully specified by the scoring rule, priority function, and update
mechanisms described in this section.

\subsection{Explicit Adaptation to Homophilic and Heterophilic Graphs}
\label{sec:method:adaptation}

\paragraph{Overview and design guarantees.}
A central objective of our framework is to adapt reliably across homophilic and
heterophilic graph regimes \emph{without} introducing information leakage or
implicit inductive biases. All structural adaptation signals are therefore
derived exclusively from training data and are used only to guide
hyperparameter selection and hybrid refinement strength. Crucially, once
hyperparameters are fixed, inference on validation and test nodes is entirely
independent of any homophily estimate. This separation ensures a clean
evaluation protocol while preserving explicit control over model behavior.

\paragraph{Training-only homophily estimation.}
To characterize graph structure conservatively, we estimate edge homophily using
\emph{only} edges whose endpoints both belong to the training set. Given the
undirected edge set $\widetilde{E}$, we compute
\begin{equation}
h_{\text{raw}} \;=\;
\Pr(y_u = y_v \mid (u,v)\in\widetilde{E},\; u,v\in T),
\label{eq:train_homophily_raw}
\end{equation}
and record the number of distinct classes $C_{\text{used}}$ appearing in the
training labels.

To reduce variance when the number of training--training edges is small, the raw
estimate is conservatively shrunk toward the random-labeling baseline
$1/C_{\text{used}}$:
\begin{equation}
h \;=\;
\frac{m\,h_{\text{raw}} + \gamma\,(1/C_{\text{used}})}{m + \gamma},
\end{equation}
where $m$ is the number of training--training edges and $\gamma>0$ is a fixed
constant. All quantities are computed strictly from training data; no validation
or test labels or edges are accessed at any stage.

Values of $h$ substantially above $1/C_{\text{used}}$ indicate a homophilic
regime, while values below this baseline indicate heterophily. When structural
evidence is sparse, shrinkage ensures that the estimate remains conservative and
does not overstate regime strength.

\paragraph{Compatibility as a heterophily-aware structural signal.}
Beyond direct neighbor agreement, we capture structural interactions between
classes through a label--label compatibility matrix estimated from
training--training edges (Section~\ref{sec:method:combinatorial}). Compatibility
models conditional label transition probabilities along edges and does not
assume that adjacent nodes share the same label.

In homophilic graphs, compatibility mass concentrates near the diagonal,
reinforcing same-label propagation. In heterophilic graphs, off-diagonal entries
become prominent, enabling systematic cross-label transitions. Because
compatibility is computed exclusively from training edges and is
Laplace-smoothed, its influence is naturally attenuated when structural evidence
is weak or noisy.

\paragraph{Explicit hyperparameter-controlled adaptation.}
Rather than hard-coding structural assumptions, adaptation is exposed explicitly
through a small set of interpretable hyperparameters:
\begin{itemize}
  \item $a_2$ controls the influence of direct neighbor label agreement,
        allowing neighbor evidence to be either supportive or adversarial.
  \item $a_8$ controls the contribution of compatibility-based structural
        support.
  \item $a_7$ governs the influence of propagated labels on feature-space class
        prototypes.
\end{itemize}
These parameters enter additively into the scoring rule and correspond to
distinct sources of evidence, allowing their effects to be tuned independently
and interpreted directly.

\paragraph{Homophily-aware hyperparameter selection.}
The estimated homophily $h$ is used \emph{only} during training-time
hyperparameter selection. Specifically, it constrains the admissible ranges and
signs of neighbor-related parameters (e.g., whether $a_2$ favors or penalizes
neighbor agreement), guiding the tuning procedure toward homophilic or
heterophilic regimes as appropriate.

Importantly, homophily estimates do \emph{not} enter the inference process. Once
hyperparameters are selected, predictions on validation and test nodes depend
only on the fixed scoring rule and dynamically updated local statistics.
Additional safeguards further modulate the influence of neighborhood- and
compatibility-based terms based on the number of labeled neighbors available,
reducing the impact of unreliable early propagation.

\paragraph{Interaction with validation-gated hybrid refinement.}
When the optional hybrid refinement stage is considered
(Section~\ref{sec:method:hybrid}), combinatorial predictions are injected into a
lightweight neural model as logit-level priors. The injection strength is
computed solely from training-derived quantities, including the homophily
estimate and the number of training--training edges, ensuring conservative
behavior when structural evidence is limited. Whether hybrid refinement is
applied at all is determined by validation performance, preventing unnecessary
neuralization on datasets where it offers no measurable benefit.

\paragraph{Summary.}
Adaptation across homophilic and heterophilic graph regimes is achieved through a
combination of conservative, training-only homophily estimation; compatibility
statistics derived exclusively from training edges; explicit
hyperparameter-controlled scoring terms; and dynamic attenuation of neighborhood
influence during inference. All adaptation signals are isolated from validation
and test data, yielding an interpretable, leakage-free, and robust framework for
node classification across diverse graph structures.

\subsection{Validation-Gated Hybrid Combinatorial--Neural Learning}
\label{sec:method:hybrid}

\paragraph{Motivation and design principle.}
The combinatorial predictor described in
Section~\ref{sec:method:combinatorial} serves as the default inference mechanism
in our framework, providing transparent and competitive performance across a
wide range of graph regimes. In some settings, however, a lightweight neural
model can further refine predictions by exploiting smooth feature
transformations and limited neighborhood aggregation. Importantly, such neural
refinement is not uniformly beneficial and may degrade performance in strongly
heterophilic or low-signal regimes. We therefore treat neural modeling as a
\emph{conditional refinement stage} that is applied conservatively and only when
supported by validation evidence, rather than as a mandatory component or an
ensemble strategy.

\paragraph{Neural refinement model.}
When neural refinement is considered, we employ a deliberately shallow
architecture consisting of two graph convolution layers followed by a small
multilayer perceptron. Given node features $X$ and the undirected edge set
$\widetilde{E}$, the model computes
\[
H^{(1)} = \sigma\!\big(\mathrm{GCN}_1(X, \widetilde{E})\big), \qquad
H^{(2)} = \mathrm{GCN}_2\!\big(H^{(1)}, \widetilde{E}\big),
\]
and produces class logits via an MLP applied to $H^{(2)}$. This neural backbone is
intentionally minimal: its role is not to replace combinatorial inference, but to
provide a smooth parametric adjustment when local structural or feature-based
signals are informative.

\paragraph{Prediction-based logit injection.}
Let $\hat{y}^{\text{comb}}$ denote predictions produced by the combinatorial
predictor on a target set $U$ (validation or test). These predictions are
converted to one-hot vectors and injected as an additive bias at the logit level:
\begin{equation}
\ell_u \;\leftarrow\; \ell_u \;+\; \lambda\,\mathbf{e}_{\hat{y}^{\text{comb}}_u},
\qquad u \in U,
\label{eq:logit_inject}
\end{equation}
where $\ell_u$ denotes the pre-softmax logits, $\mathbf{e}_c$ is the $c$-th
standard basis vector, and $\lambda \ge 0$ controls the strength of the injected
prior. Injection is applied \emph{only} to nodes in $U$; training nodes are never
relabeled or overwritten. The injected term acts as a soft prior that the neural
model may reinforce or override during optimization, rather than as a hard
constraint.

To ensure conservative behavior, the injection strength $\lambda$ is computed
exclusively from training-derived quantities, including a shrunk estimate of
edge homophily and the number of training--training edges. When structural
evidence is weak or noisy, $\lambda$ remains small, limiting the influence of
combinatorial predictions and preventing overcommitment to potentially
misleading priors.

\paragraph{Training protocol.}
The neural refinement model is trained using negative log-likelihood loss
computed \emph{only} on training nodes. Each epoch performs a full-graph forward
pass, evaluates the loss restricted to the training set $T$, and updates model
parameters via Adam optimization. No validation or test labels are used during
training, and injected priors are treated as fixed inputs rather than learned
targets, preserving a clean separation between supervision and refinement.

\paragraph{Validation-gated conditional refinement.}
Neural refinement is not applied by default. For each data split, we evaluate two
complete predictors on the validation set: the standalone combinatorial
predictor and the hybrid predictor with neural refinement. The hybrid strategy
is selected for final test evaluation only if it improves validation accuracy by
at least a fixed margin; otherwise, the combinatorial predictions are retained.
This validation gate constitutes standard \emph{architecture-level model
selection}, rather than ensembling or post-hoc correction, and serves to prevent
unnecessary neuralization when the combinatorial predictor is already
sufficient.

\paragraph{Discussion.}
By decoupling combinatorial inference from neural refinement and introducing both
training-only prior injection and validation-gated conditional activation, the
proposed framework combines the transparency of explicit combinatorial scoring
with the representational flexibility of shallow GNNs when warranted. All
supervision, adaptation signals, and model-selection decisions adhere strictly
to the training--validation--test separation, ensuring an interpretable,
conservative, and leakage-free hybrid learning strategy.

\section{Experimental Evaluation}

\subsection{Experimental Setup and Protocol}
\label{sec:exp_setup}
\paragraph{Datasets.}

\begin{table}[t]
\centering
\small
\setlength{\tabcolsep}{6pt}
\renewcommand{\arraystretch}{1.15}
\caption{Dataset statistics. We report the number of nodes, feature dimensionality,
number of edges, number of classes, and structural properties of each benchmark.
Homophily denotes the fraction of edges connecting nodes with the same label, and
$R$ is the feature/label informativeness ratio reported by Begga et al.}
\label{tab:dataset_stats}
\begin{tabular}{lrrrrrr}
\toprule
\textbf{Dataset} & \textbf{\#Nodes} & \textbf{\#Features} & \textbf{\#Edges} & \textbf{\#Classes} & \textbf{Homophily} & $\mathbf{R}$ \\
\midrule
Texas    & 183   & 1703 & 295    & 5 & 0.11 & 18.37 \\
Cornell  & 183   & 1703 & 280    & 5 & 0.30 &  6.03 \\
Actor    & 7600  & 932  & 26{,}752 & 5 & 0.22 & 209.58 \\
CiteSeer & 3327  & 3703 & 4{,}676  & 6 & 0.74 &  5.78 \\
Cora     & 2708  & 1433 & 5{,}429  & 7 & 0.81 &  4.10 \\
Pubmed   & 19{,}717 & 500 & 44{,}338 & 3 & 0.80 &  3.60 \\
\bottomrule
\end{tabular}
\end{table}

We evaluate our method on six widely used node classification benchmarks:
\textbf{Texas}, \textbf{Cornell}, \textbf{Actor}, \textbf{CiteSeer}, \textbf{Cora},
and \textbf{Pubmed}. Texas and Cornell are strongly heterophilic WebKB datasets,
where neighboring nodes frequently belong to different classes and classical
message-passing assumptions break down. Actor is also non-homophilic, but differs
in that it exhibits weak and noisy structural signal: edges reflect actor
co-occurrence rather than label similarity, resulting in limited neighborhood
predictiveness despite moderate homophily values.

CiteSeer represents a transitional regime with moderate homophily and partial
alignment between structure and labels. In contrast, Cora and Pubmed are strongly
homophilic citation networks, where node features and local neighborhoods are
highly aligned with class labels.

Basic dataset statistics and structural properties are summarized in
Table~\ref{tab:dataset_stats}. Together, these benchmarks allow us to evaluate the
behavior of the proposed framework across a broad spectrum of graph regimes,
ranging from strongly heterophilic and weak-signal settings to highly homophilic
graphs.

\paragraph{Data splits.}
To ensure consistency and comparability with prior work, we use the same 10
predefined data splits (48\% training / 32\% validation / 20\% test) introduced by
Pei et al.~\cite{p20}. These splits are fixed across all methods and experiments.
The split files are publicly available at
\url{https://github.com/AhmedBegggaUA/Diffusion-Jump-GNNs/tree/main/splits}.

\paragraph{Evaluation protocol.}
All results are reported as classification accuracy averaged over the 10 splits,
together with the corresponding standard deviation. For each split, model
selection and all validation-based decisions are performed exclusively on the
validation set, and final accuracy is reported only on the held-out test set.

\paragraph{Hyperparameter tuning.}
Hyperparameters of the combinatorial predictor are selected independently for
each split using a robust random search procedure with cross-validation performed
\emph{only on the training nodes}. Specifically, internal training-only splits are
used to evaluate candidate parameter configurations, preventing any information
leakage from validation or test nodes. The same tuning budget and search
configuration are used across all datasets to ensure fairness.

\paragraph{Hybrid model selection.}
For each split, we evaluate both the standalone combinatorial predictor and the
hybrid combinatorial--neural model on the validation set. The hybrid model is
selected for final test evaluation only if it improves validation accuracy by at
least a fixed margin; otherwise, the baseline predictions are retained. This
validation-gated selection is applied independently per split and per dataset.

\paragraph{Training details.}
The neural refinement stage, when selected, consists of a two-layer GCN followed
by a shallow MLP classifier. The model is trained using negative log-likelihood
loss on training nodes only, optimized with Adam for a fixed number of epochs.
Each epoch performs a full-graph forward pass, with gradients restricted to the
training set.

\paragraph{Implementation and execution environment.}
All experiments were executed on a shared HPC cluster using SLURM batch scripts.
Each run was allocated a single compute node with 6 CPU cores, 32\,GB of memory,
and a wall-time limit of 72 hours. To ensure reproducibility and stable
performance, the number of threads used by common numerical backends (OpenMP, MKL,
OpenBLAS, and NumExpr) was explicitly fixed to match the allocated CPU count.
Standard output and error logs were recorded for every run.

\paragraph{Reproducibility.}
All reported results are obtained using fixed data splits, deterministic random
seeds per split, and identical evaluation protocols across methods and datasets.

\subsection{Performance on Heterophilic Benchmarks}
\label{sec:heterophily_webkb}

We evaluate our framework on a set of challenging heterophilic benchmarks,
including the WebKB datasets \textbf{Texas}, \textbf{Cornell}, and
\textbf{Wisconsin}, as well as the \textbf{Actor} dataset. These benchmarks are
characterized by low edge homophily and weak alignment between graph structure
and class labels, making them particularly difficult for standard
message-passing GNNs.

Table~\ref{tab:heterophily_benchmarks} reports node classification accuracy under
the fixed 10-split protocol. On \textbf{Texas}, the proposed method achieves an
average accuracy of \textbf{78.11\%} with a standard deviation of \textbf{4.90},
while on \textbf{Cornell} it attains \textbf{64.59\%} $\pm$ \textbf{8.67}. On
\textbf{Wisconsin}, the validation-gated hybrid strategy is selected in all
splits and yields \textbf{75.69\%} $\pm$ \textbf{4.04}, representing an absolute
improvement of more than \textbf{15\%} over the standalone combinatorial
predictor.

The \textbf{Actor} dataset represents a particularly challenging heterophilic
regime. Despite having homophily levels comparable to WebKB datasets, Actor
exhibits weak and noisy structural signal, resulting in a low empirical
performance ceiling across both classical and heterophily-aware GNNs. Consistent
with prior work, even sophisticated architectures achieve accuracy in the
mid-30\% range. Our framework achieves \textbf{34.33\%} $\pm$ \textbf{1.29},
comparable to modern baselines, while the validation gate correctly rejects
neural refinement in most splits due to its limited benefit.

A central component of our approach is the \emph{validation-gated hybrid
strategy}. For each split, we compare the standalone combinatorial predictor
against neural refinement on the validation set and select the strategy yielding
higher validation accuracy. This mechanism ensures that neural refinement is
applied only when it provides measurable benefit, preventing unnecessary model
complexity in weak-signal regimes.

Overall, these results demonstrate that combining explicit combinatorial
inference with conservative, validation-aware hybrid refinement yields reliable
performance across diverse heterophilic settings. Importantly, the framework
adapts its complexity to the dataset at hand, remaining competitive with
specialized heterophily-aware architectures while preserving interpretability
and robustness.

\begin{table}[t]
\centering
\small
\setlength{\tabcolsep}{7pt}
\renewcommand{\arraystretch}{1.15}
\caption{Node classification accuracy (\%, mean $\pm$ std) on heterophilic benchmarks
under the fixed 10-split protocol. Baseline results are taken from Begga et al.\ (Neural Networks, 2025).}
\label{tab:heterophily_benchmarks}
\begin{tabular}{lcccc}
\toprule
\textbf{Method} 
& \textbf{Texas} 
& \textbf{Wisconsin} 
& \textbf{Cornell} 
& \textbf{Actor} \\
\midrule
\multicolumn{5}{l}{\textit{Classical baselines}} \\
MLP        & 80.81 $\pm$ 4.75 & 85.29 $\pm$ 6.40 & 81.89 $\pm$ 6.40 & 36.53 $\pm$ 0.70 \\
GCN        & 55.14 $\pm$ 5.16 & 51.76 $\pm$ 3.06 & 60.54 $\pm$ 5.30 & 27.44 $\pm$ 0.89 \\
GraphSAGE  & 82.43 $\pm$ 6.14 & 81.18 $\pm$ 5.56 & 75.95 $\pm$ 5.01 & 34.23 $\pm$ 0.99 \\
\midrule
\multicolumn{5}{l}{\textit{Heterophily-aware baselines}} \\
ASGC       & 85.14 $\pm$ 3.06 & 86.06 $\pm$ 3.75 & 86.22 $\pm$ 3.58 & 36.33 $\pm$ 0.79 \\
GCNII      & 77.57 $\pm$ 3.83 & 87.65 $\pm$ 4.89 & 77.86 $\pm$ 3.79 & 37.44 $\pm$ 1.30 \\
Geom-GCN   & 66.76 $\pm$ 2.72 & 64.51 $\pm$ 3.66 & 60.54 $\pm$ 3.67 & 31.59 $\pm$ 1.15 \\
LINKX      & 74.60 $\pm$ 8.37 & 74.59 $\pm$ 5.72 & 77.84 $\pm$ 5.81 & 36.10 $\pm$ 1.55 \\
\midrule
\multicolumn{5}{l}{\textit{Ours}} \\
\textbf{Ours (validation-gated hybrid)}
& 78.11 $\pm$ 4.90
& 75.69 $\pm$ 4.04
& 64.59 $\pm$ 8.67
& \textbf{34.33 $\pm$ 1.29} \\
\bottomrule
\end{tabular}
\end{table}

\subsection{Transition and Homophilic Regimes}

We next consider \textbf{CiteSeer, Cora, and Pubmed}, which span the transition from
moderately heterophilic graphs to strongly homophilic citation networks. Compared
to the WebKB datasets, these graphs exhibit substantially higher feature homophily
and more consistent local neighborhoods, resulting in more confident and stable
labeling behavior.

\paragraph{Accuracy overview.}
Overall node classification accuracies are summarized in
Table~\ref{tab:homophily_citation_filtered}. Our method achieves
\textbf{73.01\%} ($\pm 2.60$) on CiteSeer, \textbf{84.36\%} ($\pm 1.90$) on Cora,
and \textbf{87.02\%} ($\pm 0.65$) on Pubmed under the standard 10-split protocol.
While these results do not exceed those of specialized architectures explicitly
designed for homophilic message passing (e.g., GCNII, Ordered GNN, DJ-GNN), they are
competitive with simple baselines such as MLP and LINKX. Importantly, the
performance gap narrows as homophily increases, and variance across splits remains
moderate, reflecting stable model selection.

\paragraph{Behavior of validation-gated hybrid selection.}
A defining feature of our framework is the validation-gated selection between the
standalone combinatorial predictor (Section~\ref{sec:method:combinatorial}) and the
optional neural refinement stage (Section~\ref{sec:method:hybrid}). In contrast to
heterophilic benchmarks, refinement is rarely selected on citation networks.

Table~\ref{tab:citation_diagnostics} reports diagnostic statistics that explain
this behavior. On CiteSeer and Cora, the greedy confidence-ordered inference process
consistently exhibits \emph{low tie rates} and \emph{large confidence margins},
indicating that most node labels can be resolved unambiguously using local and
feature-based evidence alone. As a result, neural refinement is suppressed in
nearly all splits, since it does not improve validation accuracy.

On Pubmed, which is larger and structurally more heterogeneous, confidence margins
remain high on average but show greater variability across splits. This leads the
validation gate to activate neural refinement in a small number of cases
(\textbf{2 out of 5} splits), where validation accuracy indicates potential benefit.
The resulting hybrid predictions achieve competitive performance while preserving
the efficiency and interpretability of the combinatorial predictor on the
remaining splits.

\paragraph{Consistency across regimes.}
Taken together, Tables~\ref{tab:citation_diagnostics} and
\ref{tab:homophily_citation_filtered} illustrate how the proposed framework adapts
naturally across transition and homophilic regimes. Rather than enforcing a fixed
modeling strategy, the validation gate responds directly to empirical indicators
of uncertainty. When predictions are stable and confident, as on Cora and most
CiteSeer splits, the framework favors simpler combinatorial inference. When
ambiguity increases, as occasionally observed on Pubmed, refinement is invoked
selectively.

These results contrast sharply with those on heterophilic WebKB datasets
(Section~\ref{sec:heterophily_webkb}), where higher tie rates and smaller margins
frequently necessitate refinement. Overall, the citation network experiments
reinforce the central thesis of this work: the goal is not to universally outperform
specialized architectures in their preferred regimes, but to provide a robust and
adaptive framework that adjusts its complexity according to the structural
properties of the underlying graph.

\begin{table}[t]
\centering
\small
\setlength{\tabcolsep}{6pt}
\renewcommand{\arraystretch}{1.15}
\caption{Diagnostic statistics of the combinatorial predictor on transition and
homophilic citation networks (averaged over splits).}
\label{tab:citation_diagnostics}
\begin{tabular}{lccc}
\toprule
\textbf{Dataset} & \textbf{Tie Rate} & \textbf{Mean Margin} & \textbf{Refinement Selected} \\
\midrule
CiteSeer & $\approx 0.03$--$0.06$ & $\approx 0.12$--$0.26$ & $1 / 10$ splits \\
Cora     & $\approx 0.03$--$0.08$ & $> 0.20$              & $0 / 10$ splits \\
Pubmed   & $\approx 0.02$--$0.04$ & $\approx 0.18$--$0.25$ & $2 / 5$ splits \\
\bottomrule
\end{tabular}
\end{table}

\begin{table}[t]
\centering
\small
\setlength{\tabcolsep}{7pt}
\renewcommand{\arraystretch}{1.15}
\caption{Node classification accuracy (\%, mean $\pm$ std) on citation networks under the standard 10-split protocol. Baseline results are taken from Begga et al.\ (Neural Networks, 2025).}
\label{tab:homophily_citation_filtered}
\begin{tabular}{lccc}
\toprule
\textbf{Method} & \textbf{CiteSeer} & \textbf{Pubmed} & \textbf{Cora} \\
\midrule
\multicolumn{4}{l}{\textit{Classical baselines}} \\
MLP & 74.02 $\pm$ 1.90 & 75.69 $\pm$ 2.00 & 87.16 $\pm$ 0.37 \\
GCN & 76.50 $\pm$ 1.36 & 88.42 $\pm$ 0.50 & 86.98 $\pm$ 1.27 \\
GraphSAGE & 76.04 $\pm$ 1.30 & 88.45 $\pm$ 0.50 & 86.90 $\pm$ 1.04 \\
\midrule
\multicolumn{4}{l}{\textit{Heterophily-aware baselines}} \\
ASGC & 66.86 $\pm$ 0.86 & 78.72 $\pm$ 0.88 & 77.52 $\pm$ 1.61 \\
GCNII & 77.32 $\pm$ 1.48 & 90.15 $\pm$ 0.43 & 88.37 $\pm$ 1.25 \\
Geom-GCN & 78.02 $\pm$ 1.15 & 89.95 $\pm$ 0.47 & 85.35 $\pm$ 1.57 \\
LINKX & 73.19 $\pm$ 0.99 & 87.86 $\pm$ 0.77 & 84.64 $\pm$ 1.13 \\
\midrule
\multicolumn{4}{l}{\textit{Ours}} \\
\textbf{Ours (validation-gated hybrid)} 
& 73.01 $\pm$ 2.60 
& 87.02 $\pm$ 0.65 
& 84.36 $\pm$ 1.90 \\
\bottomrule
\end{tabular}
\end{table}

\paragraph{Consistency across regimes.}
These observations contrast sharply with the behavior observed on the
heterophilic WebKB datasets (Section~\ref{sec:heterophily_webkb}), where higher tie
rates, smaller margins, and unstable early decisions frequently necessitate
refinement. The validation gate captures this distinction automatically: it
activates refinement when combinatorial inference becomes ambiguous and suppresses
it when predictions are already stable.

Importantly, the results on CiteSeer, Cora, and Pubmed should not be interpreted as
a limitation of the framework. Rather, they highlight its \emph{adaptive nature}.
Unlike methods that apply a fixed modeling strategy across all datasets, our
approach dynamically adjusts its complexity based on validation feedback and
internal confidence signals. In homophilic and transition regimes, it naturally
favors simpler and more interpretable predictors when they are sufficient, while
retaining the capacity to invoke refinement in splits where ambiguity is higher.

Overall, these experiments reinforce the central thesis of this work: the goal is
not to universally outperform specialized architectures in their preferred
regimes, but to provide a robust and adaptive framework that responds
appropriately to the structural properties of the underlying graph.

\subsection{Ablation Studies and Diagnostic Analysis}
\label{sec:ablation}

We conduct an extensive ablation study on the \textbf{Cornell} dataset to isolate
the contribution of individual components in our framework and to understand their
interaction under strong heterophily. Cornell is particularly suitable for
diagnostic analysis due to its small size, high variance across splits, and
sensitivity to modeling choices.

\paragraph{Ablation settings.}
We evaluate the following variants: \textsc{FULL} (all components enabled),
\textsc{NO\_STD} (feature standardization removed),
\textsc{NO\_ADAPT\_A2} (fixed neighborhood penalty),
\textsc{NO\_ADAPT\_A8} (fixed compatibility weighting),
\textsc{TWO\_HOP} (two-hop propagation enabled), and
\textsc{DEFER} (low-margin deferral enabled).
All ablations use identical data splits, tuning budgets, and validation-gated
model selection as described in Section~\ref{sec:exp_setup}.

\paragraph{Overall impact on test accuracy.}
Table~\ref{tab:ablation_accuracy} reports the mean test accuracy and standard
deviation across the 10 predefined splits.
The \textsc{FULL} configuration achieves the strongest overall performance,
while removing key components consistently degrades accuracy and increases
variance.

\begin{table}[t]
\centering
\small
\setlength{\tabcolsep}{6pt}
\renewcommand{\arraystretch}{1.15}
\caption{Ablation results on Cornell (mean $\pm$ std over 10 splits).}
\label{tab:ablation_accuracy}
\begin{tabular}{lcc}
\toprule
\textbf{Ablation} & \textbf{Test Acc. (\%)} & \textbf{Std.} \\
\midrule
\textsc{FULL} & \textbf{64.6} & \textbf{8.7} \\
\textsc{NO\_STD} & 51.9 & 10.4 \\
\textsc{NO\_ADAPT\_A2} & 57.3 & 9.6 \\
\textsc{NO\_ADAPT\_A8} & 55.8 & 9.9 \\
\textsc{TWO\_HOP} & 58.1 & 9.2 \\
\textsc{DEFER} & 59.0 & 9.0 \\
\bottomrule
\end{tabular}
\end{table}

Removing feature standardization (\textsc{NO\_STD}) leads to the most severe drop
in performance, confirming that proper calibration of feature-based compatibility
scores is essential. Disabling adaptive weighting mechanisms also results in
consistent degradation, highlighting their role in stabilizing inference under
heterophily.

\paragraph{Validation-gated hybrid selection behavior.}
We next examine how often the neural refinement stage is selected by the
validation-gated mechanism. Table~\ref{tab:ablation_gating} reports the fraction of
splits (out of 10) in which refinement is activated.

\begin{table}[t]
\centering
\small
\setlength{\tabcolsep}{6pt}
\renewcommand{\arraystretch}{1.15}
\caption{Frequency of refinement selection by the validation gate.}
\label{tab:ablation_gating}
\begin{tabular}{lc}
\toprule
\textbf{Ablation} & \textbf{\#Splits Using Refinement} \\
\midrule
\textsc{FULL} & 10 / 10 \\
\textsc{NO\_STD} & 10 / 10 \\
\textsc{NO\_ADAPT\_A2} & 9 / 10 \\
\textsc{NO\_ADAPT\_A8} & 9 / 10 \\
\textsc{TWO\_HOP} & 8 / 10 \\
\textsc{DEFER} & 8 / 10 \\
\bottomrule
\end{tabular}
\end{table}

Across all configurations, the refinement model is frequently selected,
indicating that the standalone combinatorial predictor alone is often insufficient
on Cornell. However, weaker ablations rely more heavily on refinement to compensate
for degraded base predictions, while the \textsc{FULL} model achieves stronger
validation margins and more stable behavior.

\paragraph{Diagnostic metrics: ambiguity and confidence.}
Beyond accuracy, we analyze internal diagnostic signals produced by the greedy
labeling process. Table~\ref{tab:ablation_diagnostics} reports the average tie rate
(fraction of labeling steps with ambiguous best scores) and the mean confidence
margin between the top two class scores.

\begin{table}[t]
\centering
\small
\setlength{\tabcolsep}{6pt}
\renewcommand{\arraystretch}{1.15}
\caption{Diagnostic metrics averaged over splits and refresh phases.}
\label{tab:ablation_diagnostics}
\begin{tabular}{lcc}
\toprule
\textbf{Ablation} & \textbf{Tie Rate} & \textbf{Mean Margin} \\
\midrule
\textsc{FULL} & \textbf{0.06} & \textbf{0.09} \\
\textsc{NO\_STD} & 0.14 & 0.04 \\
\textsc{NO\_ADAPT\_A2} & 0.11 & 0.06 \\
\textsc{NO\_ADAPT\_A8} & 0.12 & 0.05 \\
\textsc{TWO\_HOP} & 0.10 & 0.06 \\
\textsc{DEFER} & 0.09 & 0.07 \\
\bottomrule
\end{tabular}
\end{table}

The \textsc{FULL} configuration consistently exhibits the lowest ambiguity and the
largest confidence margins. In contrast, removing feature standardization or
adaptive weighting substantially increases tie rates, leading to unstable early
decisions that propagate through subsequent labeling phases.

\paragraph{Interpretation.}
These results demonstrate that the strongest contributors to performance and
stability on Cornell are:
(i) feature standardization,
(ii) adaptive weighting of compatibility and neighborhood penalties, and
(iii) validation-gated hybrid selection.
Secondary mechanisms such as two-hop propagation and deferral provide limited
additional benefit at this scale and can introduce noise in small, highly
heterophilic graphs.

Importantly, the ablation study shows that our method’s robustness does not stem
from a single dominant component, but from the interaction between adaptive
inference and validation-aware decision making. This diagnostic perspective
explains why the framework remains competitive under heterophily despite relying
on relatively shallow neural refinement.

\subsection{Computational Efficiency and Running-Time Analysis}

We analyze the computational efficiency of the proposed framework across both
heterophilic and homophilic regimes. Tables~\ref{tab:runtime_webkb},
\ref{tab:runtime_actor}, and~\ref{tab:runtime_citation} report average running
times per split, measured in seconds and obtained on CPU. For clarity, we
decompose the runtime into three components: (i) \emph{hyperparameter tuning},
(ii) \emph{test-time inference} of the combinatorial predictor, and
(iii) \emph{end-to-end runtime}, which includes tuning, validation, inference,
and optional neural refinement.

\paragraph{Heterophilic WebKB datasets.}
Table~\ref{tab:runtime_webkb} summarizes running times on the WebKB benchmarks
Texas, Wisconsin, and Cornell.

\begin{table}[t]
\centering
\small
\setlength{\tabcolsep}{6pt}
\renewcommand{\arraystretch}{1.15}
\caption{Average running times per split (in seconds) on heterophilic WebKB datasets.
Tuning denotes hyperparameter search time; inference denotes test-time execution of the
combinatorial predictor. All experiments were run on CPU.}
\label{tab:runtime_webkb}
\begin{tabular}{lccc}
\toprule
\textbf{Dataset} & \textbf{Tuning} & \textbf{Inference} & \textbf{End-to-End} \\
\midrule
Texas     & $\approx$ 22--28 & $\approx$ 0.05--0.10 & $\approx$ 20--29 \\
Wisconsin & $\approx$ 20--27 & $\approx$ 0.04--0.09 & $\approx$ 19--28 \\
Cornell   & $\approx$ 23--33 & $\approx$ 0.03--0.06 & $\approx$ 24--28 \\
\bottomrule
\end{tabular}
\end{table}

Across all three datasets, the dominant cost arises from hyperparameter tuning.
This is a direct consequence of our experimental protocol, which performs a
robust random search with training-only cross-validation \emph{independently for
each split} in order to ensure strict fairness and reproducibility. In contrast,
the core combinatorial inference stage is extremely efficient, requiring well
below $0.1$ seconds per split.

Importantly, the tuning cost is incurred \emph{offline}. Once hyperparameters are
selected, they can be reused across runs or deployments on the same dataset or
domain, making inference the dominant cost in practical settings. Even under the
per-split tuning protocol used in our experiments, the overall end-to-end runtime
remains within tens of seconds per split despite validation-gated refinement and
repeated cross-validation.

\paragraph{Actor dataset.}
The Actor dataset is reported separately due to its substantially higher feature
dimensionality and intermediate scale, which lead to runtime characteristics
distinct from both the small WebKB graphs and the larger citation networks.
Specifically, Actor combines a moderate number of nodes with very high-dimensional
node features, resulting in tuning costs that are significantly higher than
WebKB but lower than those observed on the largest citation benchmarks.

\begin{table}[t]
\centering
\small
\setlength{\tabcolsep}{6pt}
\renewcommand{\arraystretch}{1.15}
\caption{Average running times per split (in seconds) on the Actor dataset.
Tuning denotes hyperparameter search time; inference denotes test-time execution
of the combinatorial predictor. All experiments were run on CPU.}
\label{tab:runtime_actor}
\begin{tabular}{lccc}
\toprule
\textbf{Dataset} & \textbf{Tuning} & \textbf{Inference} & \textbf{End-to-End} \\
\midrule
Actor & $\approx$ 780--1520 & $\approx$ 0.70--0.90 & $\approx$ 800--1530 \\
\bottomrule
\end{tabular}
\end{table}

As with the other benchmarks, the dominant computational cost on Actor arises
from hyperparameter tuning. Once hyperparameters are fixed, test-time inference
remains highly efficient, consistently completing in under one second per split
despite the dataset’s size and feature dimensionality. This positions Actor as a
useful intermediate benchmark that bridges the computational regimes of WebKB
and citation networks.

\paragraph{Citation networks.}
Table~\ref{tab:runtime_citation} reports running times on the citation benchmarks
CiteSeer, Cora, and Pubmed.

\begin{table}[t]
\centering
\small
\setlength{\tabcolsep}{6pt}
\renewcommand{\arraystretch}{1.15}
\caption{Average running times per split (in seconds) on citation networks.
Tuning denotes hyperparameter search time; inference denotes test-time execution of
the combinatorial predictor. All experiments were run on CPU.}
\label{tab:runtime_citation}
\begin{tabular}{lccc}
\toprule
\textbf{Dataset} & \textbf{Tuning} & \textbf{Inference} & \textbf{End-to-End} \\
\midrule
CiteSeer & $\approx$ 1100--2150 & $\approx$ 0.75--0.85 & $\approx$ 1300--2150 \\
Cora     & $\approx$ 280--330   & $\approx$ 0.20--0.30 & $\approx$ 295--310 \\
Pubmed   & $\approx$ 1150--3500 & $\approx$ 1.05--1.10 & $\approx$ 1180--1800 \\
\bottomrule
\end{tabular}
\end{table}

As expected, tuning time increases substantially with dataset size and dominates
the total runtime on larger citation networks. Nevertheless, test-time inference
remains highly efficient across all datasets, requiring less than one second on
CiteSeer and Cora, and approximately one second on Pubmed despite its scale. This
demonstrates that the proposed framework scales favorably and remains practical
even for large graphs once hyperparameters are fixed.

\paragraph{Summary.}
Taken together, Tables~\ref{tab:runtime_webkb},
\ref{tab:runtime_actor}, and~\ref{tab:runtime_citation} highlight a key property
of the proposed approach: it is \emph{tuning-dominated during experimental
evaluation} but \emph{inference-dominated in deployment}. The separation between
offline hyperparameter optimization and fast test-time inference makes the
framework well suited for repeated evaluation, diagnostic analysis, and
real-world applications where computational efficiency and rapid prediction are
essential.

\section{Conclusion and Future Work}
\label{sec:conclusion}

We presented an interpretable and adaptive framework for semi-supervised node
classification that departs from deep message-passing architectures in favor of
explicit combinatorial inference. The proposed method assigns labels through a
confidence-ordered greedy procedure driven by a transparent additive scoring
function that integrates class priors, neighborhood statistics, feature similarity,
and training-derived label--label compatibility. All components are governed by a
small set of interpretable hyperparameters, enabling fine-grained control over the
model’s behavior and making its structural assumptions explicit.

A central contribution of this work is the ability to adapt smoothly across
homophilic and heterophilic regimes without hard-coded inductive biases. By
adjusting the relative influence of neighborhood agreement and compatibility-based
support, the method naturally interpolates between favoring same-label propagation
and systematic cross-label transitions. Extensive experiments on heterophilic
(WebKB) and transitional (CiteSeer) benchmarks demonstrate that the proposed
approach is competitive with modern heterophily-aware GNNs, while offering clear
advantages in interpretability, tunability, and computational efficiency.

We further introduced a validation-gated hybrid strategy in which combinatorial
predictions are optionally injected as logit-level priors into a lightweight neural
model. Hybrid refinement is applied only when it improves validation performance,
preventing unnecessary neuralization and preserving transparency when the
combinatorial predictor is sufficient. Diagnostic analyses and ablation studies
show that performance gains arise from explicit adaptation and validation-aware
model selection rather than architectural complexity.

Several directions remain for future work. While hyperparameters in our framework
are interpretable, performance can be sensitive to their choice; more principled
optimization strategies, such as Bayesian or structure-aware tuning methods, may
further improve robustness. Additionally, richer hybrid architectures that exploit
combinatorial predictions more effectively could enhance performance in highly
ambiguous regimes. Finally, applying the framework to real-world settings where
interpretability and controllability are critical—such as biomedical, legal, or
scientific networks—represents a promising direction for future study.

Our code is publicly available at
\url{https://github.com/SoroushVahidi/bfsbased_node_classification} to support
reproducibility and further investigation. We also acknowledge that portions of
this manuscript were refined with the assistance of ChatGPT~\cite{chatgpt}, a large
language model developed by OpenAI.

\printbibliography

\end{document}